\documentclass[10pt, logo, copyright]{nvidiatechreport}

\usepackage{mdframed}
\usepackage[utf8]{inputenc} 
\usepackage[T1]{fontenc}    

\usepackage{amsfonts}       
\usepackage{nicefrac}       
\usepackage{microtype}      

\usepackage[numbers]{natbib}
\usepackage{tabto}
\usepackage{xspace}
\usepackage{adjustbox}
\usepackage{wrapfig}
\usepackage{dblfloatfix}
\usepackage{footmisc}
\usepackage{float}
\usepackage{listings}

\usepackage{natbib}
\usepackage{hyperref}
\usepackage{url}
\usepackage{booktabs}
\usepackage{multirow}
\usepackage{tabularx}
\usepackage{arydshln}

\usepackage{colortbl}
\usepackage{amsmath}
\usepackage{makecell}
\usepackage{hhline}
\usepackage{array}
\usepackage{diagbox}
\usepackage{graphicx}

\usepackage{fp} 

\usepackage{bbm}
\usepackage{amsmath}
\usepackage{amssymb}
\usepackage{fontawesome}
\usepackage{pifont}
\usepackage{MnSymbol}
\usepackage[ruled,lined]{algorithm2e}
\DontPrintSemicolon
\SetKwInOut{Intermediate}{Intermediate}

\usepackage{booktabs}
\usepackage{multirow}
\usepackage{wrapfig}
\usepackage{makecell}
\usepackage{rotating}
\definecolor{periwinkle}{HTML}{7977B8}

\usepackage{hyperref}
\usepackage{url}
\usepackage{titlesec}
\usepackage[capitalize,noabbrev]{cleveref} 
\usepackage{enumitem}

\crefname{problem}{problem}{problems}
\crefname{aview}{alternative view}{alternative views}
\crefname{argument}{argument}{arguments}
\crefname{cargument}{counter-argument}{counter-arguments}
\crefname{step}{step}{steps}
\crefname{barrier}{barrier}{barriers}
\crefname{wdefinition}{definition}{definitions}

\title{Universal Deep Research:\\Bring Your Own Model and Strategy}

\author{
Peter Belcak,
Pavlo Molchanov
\\
NVIDIA Research
}

\titlespacing*{\paragraph}
  {0pt}   
  {0pt}   
  {6pt}   

\makeatletter
\def\gcmidrule{\arrayrulecolor{lightgray}
    \noalign{\ifnum0=`}\fi
    \@ifnextchar[{\@gcmidrule}{\@gcmidrule[\cmidrulewidth]}}
\def\@gcmidrule[#1]{\@ifnextchar({\@@gcmidrule[#1]}{\@@gcmidrule[#1]()}}
\def\@@gcmidrule[#1](#2)#3{\@@@gcmidrule[#3]{#1}{#2}}
\def\@@@gcmidrule[#1-#2]#3#4{\global\@cmidla#1\relax
    \global\advance\@cmidla\m@ne
    \ifnum\@cmidla>0\global\let\@gtempa\@cmidrulea\else
    \global\let\@gtempa\@cmidruleb\fi
    \global\@cmidlb#2\relax
    \global\advance\@cmidlb-\@cmidla
    \global\@thisrulewidth=#3
    \@setrulekerning{#4}
    \ifnum\@lastruleclass=\z@\vskip \aboverulesep\fi
    \ifnum0=`{\fi}\@gtempa
    \noalign{\ifnum0=`}\fi\futurenonspacelet\@tempa\@xgcmidrule}
\def\@xgcmidrule{%
   \ifx\@tempa\gcmidrule
       \vskip-\@thisrulewidth
       \global\@lastruleclass=\@ne
   \else \ifx\@tempa\morecmidrules
       \vskip \cmidrulesep
       \global\@lastruleclass=\@ne\else
       \vskip \belowrulesep
       \global\@lastruleclass=\z@
   \fi\fi
   \ifnum0=`{\fi}
  \arrayrulecolor{black}}
\makeatother

\correspondingauthor{X}

\begin{abstract}
\end{abstract}

\begin{document}
\maketitle

\textbf{Abstract.}

Deep research tools are among the most impactful and most commonly encountered agentic systems today. We observe, however, that each deep research agent introduced so far is hard-coded to carry out a particular research strategy using a fixed choice of tools.

We introduce Universal Deep Research (UDR), a generalist agentic system that wraps around any language model and enables the user to create, edit, and refine their own entirely custom deep research strategies without any need for additional training or finetuning.

To showcase the generality of our system, we equip UDR with example minimal, expansive, and intensive research strategies, and provide a user interface to facilitate experimentation with the system.

\textbf{Links:} 
\href{https://research.nvidia.com/labs/lpr/udr}{Project} | 
\href{https://github.com/NVlabs/UniversalDeepResearch}{Code} | 
\href{https://nv-dler.github.io}{Lab}

\section{Introduction}
\label{section:introduction}
Deep research tools are a recently emerged but already popular class of instruments for carrying out search-intensive tasks in many white-collar professions.
In private use, they serve as useful gadgets for continued learning and the satisfaction of personal curiosity.

The function of a deep research tool (DRT) is to take a \textit{research prompt} from the user, conduct an extensive search of the available resources relevant to the task specified in the prompt, and produce a \textit{research report} that is structured and formatted according to the requirements specified in the prompt.

\begin{figure}[b!]
    \centering
    \includegraphics[width=0.98\linewidth]{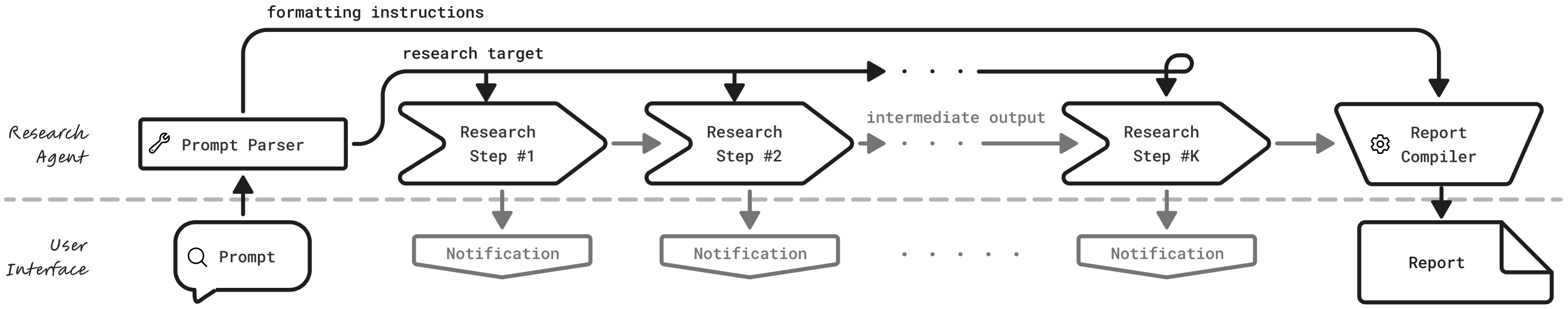}
    \caption{A high-level diagram visualizing the components of a typical deep research tool. Unlike plain conversational LLMs, DRTs tend to continuously update the user on their progress before producing their report.}
    \label{figure:drt}
\end{figure}

A DRT consists of
\begin{enumerate}[label={(\alph*)}]
    \item a simple user interface designed to receive the research prompt, continuously update the user on the progress of the research, and to display the research report; and
    \item agentic logic, implemented either through code agency as a code-orchestrated use of LLMs and tools or via LLM agency leveraging reasoning and model tool-calling \cite{belcak2025small}.
\end{enumerate}
This is illustrated in \Cref{figure:drt}.
The research report produced by a DRT typically contains structural elements such as headings and tables, extensive formatting, and relevant references to the sources used in producing the report, which sets it apart from the more compact responses seen in conversational agents \cite{masterman2024landscape}.

\paragraph{General landscape.}
A number of DRTs exist and are under active development.
Gemini, Perplexity, and OpenAI Deep Research tools \cite{citron2024DeepResearch,Perplexity2025DeepResearch,OpenAI2025DeepResearch} transform user prompts into a research plan and analyze the web by browsing autonomously, at each step finding information and then starting new searches based on the results of previous searches. This iterative process continues until the DRT deems it has gathered sufficient data to generate a comprehensive report.
Grok 3 DeepSearch \cite{Zhou2025UnderstandingGrok} employs a two-tier crawling architecture, in which a distributed network of crawler bots continuously indexes the web. When a user submits a query, the DRT activates an on-demand agent that performs targeted searches by generating specific sub-queries, fetching relevant pages in real-time. The system additionally leverages chain-of-thought LM reasoning to process retrieved data, evaluating source credibility and cross-verifying claims across multiple sources.

\paragraph{Enterprise landscape.}
While iterative (``intensive'') and wide-spanning (``expansive'') research strategies appear to function well where the space of acceptable sources is wide (e.g., the entirety of the web as in the case of the above tools), the enterprise-focused DRTs tailored to perform research inside the much more limited internal document databases appear to employ considerably more specialized approaches.
The NVIDIA AI-Q Research Assistant \cite{NVIDIAAIQResearchAssistant2025} follows a structured prompt-focused five-step process: (1) creating a report plan based on the research prompt, (2) searching data sources for answers, (3) writing a report, (4) reflecting on gaps in the report for further queries if needed, and (5) finishing with a comprehensive list of document sources.
SambaNova Deep Research \cite{sambanova2025deepresearch} similarly operates through a document-oriented five-step pipeline: (I) parsing the prompt to identify the research scope, (II) generating a detailed document outline with section-level planning (cf. AI-Q with prompt-level planning), (III) gathering information using web tools and APIs across sources, (IV) delegating tasks to specialized research agents such as the financial agent for hierarchical analysis, and (V) synthesizing findings into Markdown-formatted reports with citations.
Perhaps the most idiosyncratic of the tools surveyed, ERP AI Deep Research \cite{ERP_AI_DeepResearch} employs a ``Graph-Based AI Architecture'' that represents enterprise data through knowledge graphs and accesses them using Graph Neural Networks. Being fundamentally graph-oriented, it does not follow the traditional document/page notion of sources and performs rounds of information aggregations across relevant graph elements rather than document web traversals combined with LLM text comprehension.

\paragraph{Problem statement.}
The existing DRTs employ predominantly rigid research strategies with little room for user customization beyond the research prompt, and, in case of DRTs with LLM agency, rely on a single choice of an underlying model or of a model family with identical behavioral post-training.

While this problem is not a bar to the DRTs' wide popularity, it limits their utility in three ways:
\begin{enumerate}[label={\textbf{P\arabic*}}]
    \item\label[problem]{problem:p1} The user ability to enforce a hierarchy of preferred resources, automate cross-validation of information against reputable sources, and take charge of the expenses associated with each search is restricted. This contributes to the reasons for the existence of the functionality gap between customer- and enterprise-oriented DRTs.
    \item\label[problem]{problem:p2} The creation of specialized document research strategies that are ultimately necessary in high-value industries is not possible within the existing DRTs. This leaves large parts of professional research workloads in these industries to be automated by costly agentic solutions made to measure.
    \item\label[problem]{problem:p3} The models used in the existing DRTs are not interchangeable -- one cannot take the most recent or the most powerful model and freely combine it with a deep research agent of their choice in order to yield an even more powerful DRT.
\end{enumerate}

\paragraph{Problem importance and impact.}
To resolve \Cref{problem:p1} would help to increase the quality of research reports for individual users and help to close the gap between customer and enterprise DRTs.
To resolve \Cref{problem:p2} would enable the automation of a large amounts of high-value labor-intensive specialized research workloads in industries such as finance, legal, healthcare, and real estate, or government and public administration.
To resolve \Cref{problem:p3} would allow the pairing of the most competitive models with the most competitive DRTs, and allow for unspecialized competition among models and DRTs independently of each other.

\paragraph{Contribution.}
We introduce Universal Deep Research (UDR), a generalist agentic system that wraps around any language model, does not require additional fine-tuning to function, and enables the user to create, edit, and refine their own entirely custom deep research strategies without any need for additional training or finetuning.
The key component of UDR's research mechanism is the conversion of user-defined research queries into actionable code snippets within the bounds of the permissible control flow and available tools. Being thus equipped, UDR allows the user to define complex procedures for document research, validation, and report structuring, assuming complete control of any aspect of the research process.

\paragraph{Novelty.}
The novelty of our contribution resides in it presenting a general solution to resolve \cref{problem:p1,problem:p2,problem:p3}.

\paragraph{Target audience.}
The end-to-end system demonstrating UDR in this work is a research prototype. Its target audience are developers working on agentic solutions in the industry and members of the research community investigating the advantages and limitations of agentic systems.

\section{Research Mechanism}
\label{section:udr}

UDR takes both the intended research strategy and the research prompt as inputs and follows the instructions of the strategy, abstaining from using any individual agency unless explicitly instructed to do so.

\subsection{Inputs}

\paragraph{Research Strategy.}
The research strategy defines the behavior of the UDR research instance in its entirety.
Typically, a research strategy would consist of a list of steps, preferably formatted as a numbered or bullet-point list to make it easier for the strategy compiler to process to identify and separate the individual steps.
We give examples of such research strategies \Cref{appendix:example_research_strategies}.

\paragraph{Research Prompt.}
Similarly to other DRTs, UDR accepts a research prompt that specifies what topic or question is to be researched. There are no implicit restrictions on the nature of the prompt -- if a condition is to be enforced, it must be checked and reported to the user as a part of the research strategy. A typical deep research prompt consists of a query, content requirements, and formatting requirements. It is up to the research strategy to extract all information relevant to the approach it implements from the prompt.
Examples of research prompts can be found in \Cref{appendix:example_inputs_and_outputs}.

\subsection{Operation}
\begin{figure}[b!]
    \centering
    \includegraphics[width=0.98\linewidth]{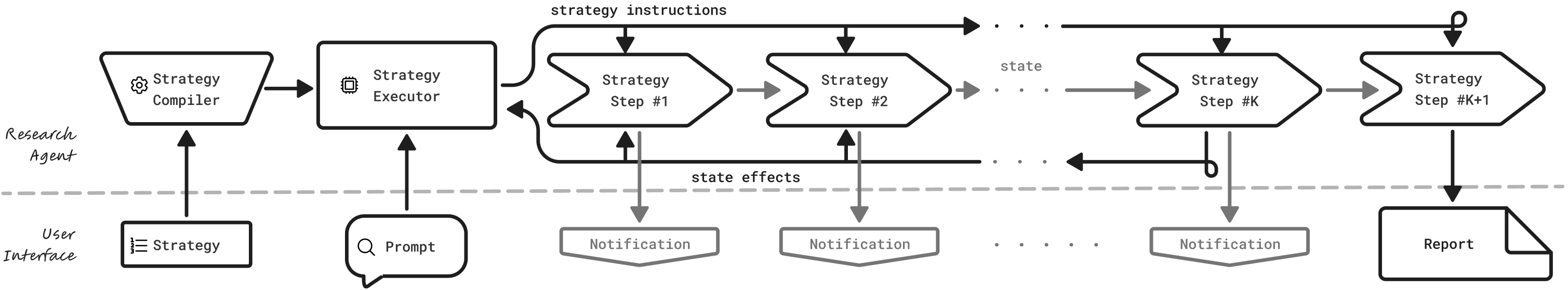}
    \caption{A high-level diagram visualizing the components of the UDR. Unlike specialized DRT, UDR receives both the research strategy and the research prompt from the user, allowing for a greater level of customization.}
    \label{figure:udr}
\end{figure}

The operation of UDR is illustrated in \Cref{figure:udr}. It consists of two major phases.

\paragraph{Phase 1 -- Strategy processing.}
The core of the UDR functionality is in its conversion of user-specified natural language research strategies into executable code built with regard to the tool use and communication mechanisms of the UDR.
Once the research strategy has been received from the user, it is passed to a language model together with constraints on the available functions and permitted code structures.
The strategy is converted to a single callable function that accepts the research prompt as the input and continuously returns output notifications. In our implementation, we enforced this behavior by insisting that the generated function returns a generator and that every notification sent to the user that updates them on the progress of the research is a \texttt{yield} statement returning a dictionary with the notification payload.
We found that regardless of the choice of the language model, giving it a completely free hand at generating the code implementation of the research strategy often resulted in the model taking shortcuts, skipping strategy steps, and taking liberties at imposing constraints where none were stipulated by the user.
To all but eradicate this behavior, we prompted the model to generate code that corresponded to the strategy step by step, explicitly prepending every segment of the generated code by comments laying out the strategy step it corresponds to.
The available tools, namely the search function, was described in a docstring contained in the user message to the code-generating model.

\paragraph{Phase 2 -- Strategy execution.}
Once the research strategy has been converted to code and its conformance with execution requirements has been confirmed, it is invoked in an isolated code execution environment.
We note a number of crucial operational details below.
\begin{itemize}
    \item \textbf{State modifications.} Rather than maintaining a single growing context window, UDR stores all intermediate information and text fragments as named variables in the code execution state. This approach enables the system to operate entirely within a small context window. In our experiments, a context length of 8k tokens was sufficient to carry out full research workflows, regardless of their complexity.

    \item \textbf{Tool use.} All tools are accessed via synchronous function calls, ensuring transparent and deterministic behavior. Because the state is maintained outside the LM context—in persistent code variables—the system can accurately refer to and reuse information gathered in earlier steps, even if that information was processed well before the current point in execution. The architecture also allows for a future upgrade to asynchronous tool use for improved performance.

    \item \textbf{LM reasoning.} Language model reasoning is treated as a callable utility rather than a controlling entity. UDR uses the language model for localized reasoning tasks such as summarization, ranking, or extraction, in line with specific steps in the user-defined research strategy. This contrasts with typical DRTs, where the LM often orchestrates the full research process.

    \item \textbf{Notifications.} Throughout the execution of a research strategy, the user is kept informed via structured notifications, which are explicitly defined by the strategy author. These progress updates are emitted using \texttt{yield} statements in the generated code, allowing the user interface to display real-time updates without revealing raw intermediate outputs or internal state unless explicitly requested.
\end{itemize}

\paragraph{Reliability.}
We found that generating code to follow the user-defined research strategy in a single end-to-end pass yielded significantly more reliable outcomes than earlier approaches. Initial attempts—such as embedding the strategy directly within the prompt to a reasoning-oriented language model, or decomposing the strategy into individual steps and generating isolated code fragments for each—proved fragile and error-prone. In contrast, the current approach ensures coherence across steps, minimizes synchronization overhead, and avoids the cascade of failures often seen in fragmented orchestration. The resulting code is fully interpretable and auditable by the user, and it rarely exhibits the failure modes encountered in our earlier prototypes, such as skipping strategy steps, misapplying search instructions, invoking the language model out of sequence, or introducing spurious checks and constraints not present in the original strategy. These improvements are attributable to the disciplined structure enforced during code generation and the contextual clarity of handling the entire strategy holistically.

\paragraph{Efficiency.}
UDR achieves high computational efficiency by separating control logic from language model reasoning. The orchestration of the deep research process is handled entirely by generated code, which executes on the CPU without requiring orders-of-magnitude more expensive language model inference. Language model calls are invoked only where explicitly instructed by the user-defined research strategy, and these calls operate on compact, well-scoped textual fragments stored in the code’s variable state. This dual-level efficiency --- delegating orchestration to CPU-executable logic and limiting LLM use to focused, context-efficient invocations --- not only reduces GPU usage but also minimizes the overall latency and cost of executing deep research tasks.

\paragraph{Security.}
Because UDR generates and executes user-defined code, it is essential to account for the risks associated with prompt injection and code-based exploits. To ensure robust isolation, each generated strategy can be, by our design, executed within a sandboxed environment that prevents access to the host system. This isolation is enforced by default and is designed to eliminate the possibility of side effects beyond the execution context. Ready-to-use solutions such as Piston \cite{man_piston_2025} provide a foundation for such execution environments. We emphasize that isolating the execution layer is a strict requirement for any deployment of UDR beyond that to a fully trusted audience.

\subsection{Outputs}

\paragraph{Notifications.}
Throughout the execution of a user-defined research strategy, UDR emits structured progress updates via \texttt{yield} statements in the generated code. Each yielded notification is a dictionary object containing a loosely pre-specified schema with fields such as \texttt{type}, \texttt{timestamp}, and \texttt{description}. These notifications are intended to be parsed and displayed in real time by the user interface, providing an interpretable, low-latency view into the ongoing execution of the research strategy. Because each notification is explicitly authored as part of the strategy, the user retains full control over the granularity and content of progress updates. This mechanism replaces the implicit, often opaque progress tracking found in typical DRTs with a deterministic, user-auditable stream of events.

\paragraph{Research Report.}
The final output of a UDR execution is a complete research report constructed by the strategy in accordance with the user's prompt and preferences. The report is returned as the final yielded notification, marked by a distinctive \texttt{type} (e.g., \texttt{"final\textunderscore{}report"}) to signal termination of the research procedure. The report may include structured text, markdown formatting, tables, references, and any other components dictated by the strategy logic. Because the report is built entirely from the accumulated variable states, it reflects the full traceability of the research process and can be verified against the inputs and tools invoked. This explicit, user-controlled (through the specification of the strategy) final-stage construction allows UDR to deliver reproducible, format-consistent outputs wherever desired by the user. Examples of research reports (and the corresponding research prompts) can be found in \Cref{appendix:example_inputs_and_outputs}.

\section{User Interface}
\label{section:ui}
UDR is, in principle and in broad terms, compatible with any existing user interface designed to perform deep research.
This is because the research strategy, despite being an additional input parameter to the system when compared to other tools, can be extracted from a research prompt with the help of a language model.

\begin{figure}[h!]
    \centering
    \includegraphics[width=0.50\linewidth]{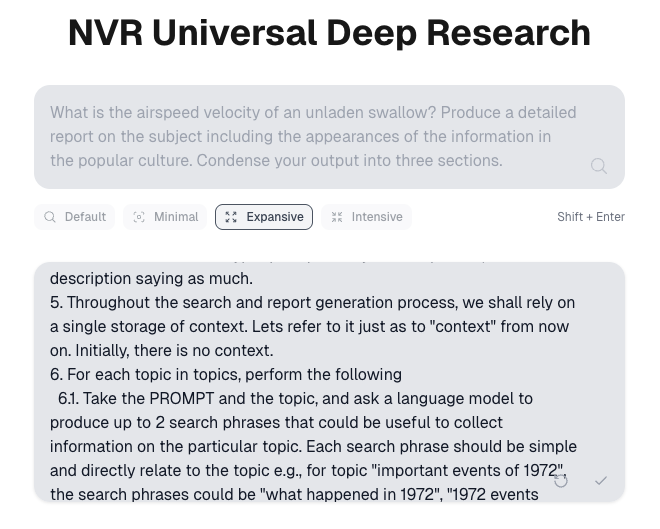}
    \caption{A screenshot of the user interface developed for the purpose of demonstrating UDR showing the search bar (\textit{top}), strategy selection list (\textit{middle}), and the strategy editing text area (\textit{bottom}).}
    \label{figure:ui1}
\end{figure}

\begin{figure}[h!]
    \centering
    \includegraphics[width=0.90\linewidth]{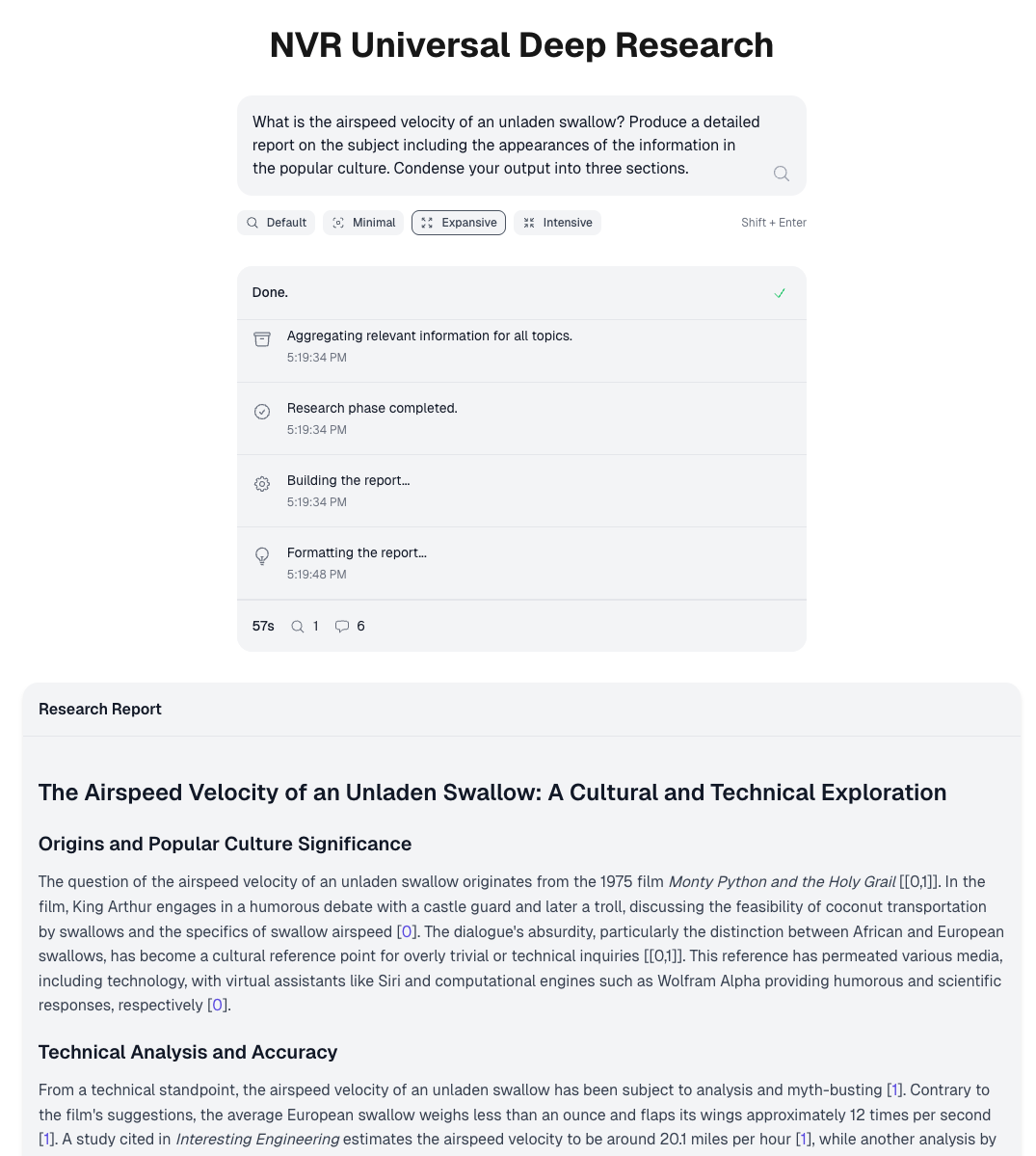}
    \caption{A screenshot of the UDR demonstration UI showing a completed research workflow, featuring the search bar (\textit{top}), strategy selection list (\textit{top-middle}), research progress notification visualizer (\textit{bottom-middle}), and the report viewer (\textit{bottom}).}
    \label{figure:ui2}
\end{figure}

However, to demonstrate the flexibility of UDR, we also developed out own demonstration user interface, pictured in \Cref{figure:ui1,figure:ui2}.

\begin{itemize}
    \item \textbf{Search bar.} This allows the user to input their research prompt, specifying the topic, questions, and desired output format for their deep research. This is the primary input mechanism for initiating a research task.

    \item \textbf{Strategy selection list.} This displays a list of pre-existing research strategies that the user has previously created or saved. Users can select a strategy from this list, enabling them to quickly apply a predefined research methodology to their current prompt without needing to recreate it.

    \item \textbf{Edit strategy button.} Clicking this button activates the strategy editing interface. It enables users to modify the currently selected research strategy, providing granular control over the research process.

    \item \textbf{Strategy editing text area.} When the ``edit strategy'' button is clicked, this text area appears, pre-populated with the details of the currently selected research strategy. Users can directly write, edit, and refine their custom research strategies here using natural language, which UDR then converts into executable code.

    \item \textbf{Research progress notifications.} This section presents a real-time, chronological list of updates and notifications from the UDR system as it executes the research strategy. Each notification includes an icon for quick visual identification, a descriptive message outlining the current status or action, and a timestamp for traceability. This feature provides transparency into the research process, allowing users to monitor progress and understand the steps being taken.

    \item \textbf{Stop research button.} This allows the user to interrupt an ongoing research process at any point. This is crucial for controlling resource usage or for stopping research that is not yielding desired results.

    \item \textbf{Generate report button.} This button becomes visible and active only if the research process has been stopped manually by the user, and if at least one research result has been processed (i.e., after the initial notification of the first result and before a final report candidate is generated). This provides the user with the flexibility to generate a preliminary research report based on the information gathered up to the point of interruption, even if the full strategy has not been completed.

    \item \textbf{Report viewer.} This dedicated area is designed to render the final research report in a user-friendly, Markdown-formatted display. It ensures that all structural elements like headings, tables, and formatting are correctly visualized, making the comprehensive research output easily digestible and reviewable by the user.
\end{itemize}

\section{Limitations}
\label{section:limitations}

\paragraph{Reliance on language model code generation.}
The faithfulness of UDR's behavior to the given research strategy depends on the quality of code generated by the underlying language model. While we mitigate inconsistency through enforced comment-code structure, occasional semantic drift or hallucinated logic may still occur, particularly with ambiguous or underspecified strategies.

\paragraph{Trust in user-defined strategies.}
UDR assumes that the research strategy authored by the user is logically sound, safe, and fit for purpose. The system does not currently validate whether the specified steps produce a coherent or meaningful workflow beyond basic syntactic and execution checks. As a result, poorly designed strategies will yield ineffective or incomplete reports, or may not return reports at all.

\paragraph{Limited real-time interactivity.}
Although UDR provides live progress notifications via structured yields, the current implementation does not support mid-execution user intervention (beyond user stopping the workflow) or dynamic branching based on real-time user feedback. All decision logic must be encoded upfront in the research strategy, limiting adaptability in long or exploratory research workflows.

\section{Conclusions \& Recommendations}
\label{section:conclusions}
UDR demonstrates that it is feasible to attach a well-functioning deep research tool on top of virtually any general-enough generative language model, that it is simultaneously possible to give the end user easily understandable agency over the strategy of the deep research process.
However, devising a research strategy sufficiently sophisticated to contend with the complexities of all user queries and topics has proven to be a rather tedious process for the end users of the applications -- even those who appreciate the level of control this approach lends them.
Furthermore, UDR shows not only that it is feasible but that it might also sometimes be desirable to allow the user to implement (``program'') agentic behavior in natural language.
We see a great level of potential utility in giving the user direct control over the agency of their language model.

Noting the above, we recommend that
\begin{enumerate}[label={\textbf{R\arabic*}}]
    \item a system similar to UDR, if deployed to the end consumer, be equipped with a library of research strategies for modification and customization rather than the requirement that the user bring their own strategy;
    \item it be further explored how to give users control over the otherwise free reasoning (sometimes also referred to as ``thinking'') of language models; and

    \item it be further explored how potentially a large set of user prompts could be automatically turned into deterministically controlled agents performing complex series of actions on their behalf.
\end{enumerate}

\clearpage
{
  \small
  \bibliographystyle{unsrt}
  \bibliography{bibliography}
}

\clearpage
\appendix

\section{Example Research Strategies}
\label{appendix:example_research_strategies}

\subsection{Minimal Research Strategy}
\label{appendix:minimal_research_strategy}
\begin{lstlisting}[breaklines]
1. Send a notification of type "prompt_received" with description saying what PROMPT has been received, e.g. "Received research request: {PROMPT}"
2. Send a notification of type "prompt_analysis_started", with description indicating that we are now analyzing the research request.
3. Take the PROMPT and ask a language model to produce 3 search phrases that could help with retrieving results from search engine for the purpose of compiling a report the user asks for in the PROMPT. The search phrases should be simple and objective, e.g. "important events 1972" or "energy consumption composition in India today". Use a long prompt for the model that describes in detail what is supposed to be performed and the expected output format. Instruct the model to return the search phrases on one line each. Tell the model not to output any other text -- just the newline-separated phrases. Then, parse the output of the language model line by line and save the resulting search phrases as "phrases" for further research, skipping over empty lines.
4. Send a notification of type "prompt_analysis_completed", with a description saying as much.
4.1 Send a notification of type "task_analysis_completed", informing the user that the search plan has been completed and informing them how many search phrases will be invoked, e.g.  "Search planning completed. Will be searching through {len(topics)}+ terms."
5. For each phrase in phrases output by step 3., perform the following:
    - Send a notification of type "search_started", with the description indicating what search phrase we are using for the search, e.g. "Searching for phrase '{phrase}'"
    - Perform search with the phrase.
    - Once the search returns some results, append their contents to CONTEXT one by one, separating them by double newlines from what is already present in the CONTEXT.
    - Send a notification of type "search_result_processing_completed", indicating in its description that the search results for term {term} have been processed.
6. Send a notification to the user with type "research_completed", indicating that the "Research phase is now completed.".
7. Send a notification with type "report_building", with the description indicating that the report is being built.
8. Take CONTEXT. Call the language model, instructing it to take CONTEXT (to be appended into the LM call) and produce a deep research report on the topic requested in PROMPT. The resulting report should go into detail wherever possible, rely only on the information available in CONTEXT, address the instruction given in the PROMPT, and be formatted in Markdown. This is to be communicated in the prompt. Do not shy away from using long, detailed and descriptive prompts! Tell the model not to output any other text, just the report. The result produced by the language model is to be called REPORT.
9. Send a notification with type "report_done", indicating that the report has been completed. Add "report" as a field containing the REPORT to be an additional payload to the notification.
10. Output the REPORT.
\end{lstlisting}

\subsection{Expansive Research Strategy}
\label{appendix:expansive_research_strategy}
\begin{lstlisting}[breaklines]
1. Send a notification of type "prompt_received" with description saying what PROMPT has been received, e.g. "Received research request: {PROMPT}"
2. Send a notification of type "prompt_analysis_started", with description indicating that we are now analyzing the research request.
3. Take the PROMPT and ask a language model to produce 2 topics that could be useful to investigate in order to produce the report requested in the PROMPT. The topics should be simple and sufficiently different from each other, e.g. "important events of 1972" or "energy consumption composition in India today". Instruct the model to return the topics on one line each. Tell the model not to output any other text. Then, parse the output of the language model line by line and save the resulting topics as "topics" for further research.
4. Send a notification of type "prompt_analysis_completed", with description saying as much.
5. Throughout the search and report generation process, we shall rely on a single storage of context. Lets refer to it just as to "context" from now on. Initially, there is no context.
6. For each topic in topics, perform the following
  6.1. Take the PROMPT and the topic, and ask a language model to produce up to 2 search phrases that could be useful to collect information on the particular topic. Each search phrase should be simple and directly relate to the topic e.g., for topic "important events of 1972", the search phrases could be "what happened in 1972", "1972 events worldwide", "important events 1971-1973". For topic "energy consumption composition in India today", the search phrases could be "renewable energy production in India today", "fossil fuel energy reliance India", "energy security India". Call the returned phrases simply "phrases" from now on.
  6.2. For each phrase in phrases output by step 6.1., perform the following:
    - Send a notification of type "search_started", with the description indicating what search phrase we are using for the search, e.g. "Searching for phrase '{phrase}'"
    - Perform search with the phrase. Once the search returns some results, append their contents to context one by one, separating them by double newlines from what is already present in the context.
    - Send a notification of type "search_result_processing_completed", indicating in its description that the search results for term {term} have been processed.
7. Send a notification with type "report_building", with the description indicating that the report is being built.
8. Take CONTEXT. Call the language model, instructing it to take context (to be appended into the LM call) and produce a deep research report on the topic requested in PROMPT. The resulting report should go into detail wherever possible, rely only on the information available in context, address the instruction given in the PROMPT, and be formatted in Markdown. This is to be communicated in the prompt. Do not shy away from using long, detailed and descriptive prompts! Tell the model not to output any other text, just the report. The result produced by the language model is to be called REPORT.
9. Send a notification with type "report_done", indicating that the report has been completed. Add "report" as a field containing the REPORT to be an additional payload to the notification.
10. Output the REPORT.
\end{lstlisting}

\subsection{Intensive Research Strategy}
\label{appendix:intensive_research_strategy}
\begin{lstlisting}[breaklines]
1. Send a notification of type "prompt_received" with description saying what PROMPT has been received, e.g. "Received research request: {PROMPT}"
2. Send a notification of type "prompt_analysis_started", with description indicating that we are now analyzing the research request.
3. Throughout the search and report generation process, we shall rely on two storages of context. One shall be called "supercontext" and contain all contexts of all resources read throughout the search phase. The other one shall be called "subcontext" and pertain to only one interation of the search process. At the beginning, both the supercontext and subcontext are empty.
4. Take the PROMPT and ask a language model to produce 2 search phrases that could help with retrieving results from search engine for the purpose of compiling a report the user asks for in the PROMPT. The search phrases should be simple and objective, e.g. "important events 1972" or "energy consumption composition in India today". Use a long prompt for the model that describes in detail what is supposed to be performed and the expected output format. Instruct the model to return the search phrases on one line each. Tell the model not to output any other text -- just the newline-separated phrases. Then, parse the output of the language model line by line and save the resulting search phrases as "phrases" for further research, skipping over empty lines.
4.1. Send a notification of type "prompt_analysis_completed", with a description saying as much.
5. Perform the following 2 times:
 - Clear the subcontext.
 - For each phrase in phrases, perform the following:
    * Send a notification of type "search_started", with the description indicating what search phrase we are using for the search, e.g. "Searching for phrase '{phrase}'"
    * Perform search with the phrase. Once the search returns some results, append their contents to subcontext one by one, separating them by double newlines from what is already present in the subcontext.
    * Send a notification of type "search_result_processing_completed", indicating in its description that the search results for term {term} have been processed.
 - Once the subcontext has been put together by aggregating the contributions due to all search phrases, ask a language model, given the subcontext and the PROMPT given by the user, to come up with 2 more phrases (distinct to phrases that are already in phrases) on the basis of the new subcontext being available. Again, the search phrases should be simple and objective, e.g. "important events 1972" or "energy consumption composition in India today". Use a long prompt for the model that describes in detail what is supposed to be performed and the expected output format. Instruct the model to return the search phrases on one line each. Tell the model not to output any other text -- just the newline-separated phrases. Then, parse the output of the language model line by line and save the resulting search phrases as "phrases" for further research, skipping over empty lines. Clear all the old phrases and let the newly returned phrases by the phrases for the next iteration of this loop.
6. Send a notification with type "report_building", with the description indicating that the report is being built.
7. Take CONTEXT. Call the language model, instructing it to take CONTEXT (to be appended into the LM call) and produce a deep research report on the topic requested in PROMPT. The resulting report should go into detail wherever possible, rely only on the information available in CONTEXT, address the instruction given in the PROMPT, and be formatted in Markdown. This is to be communicated in the prompt. Do not shy away from using long, detailed and descriptive prompts! Tell the model not to output any other text, just the report. The result produced by the language model is to be called REPORT.
8. Send a notification with type "report_done", indicating that the report has been completed. Add "report" as a field containing the REPORT to be an additional payload to the notification.
9. Output the REPORT.
\end{lstlisting}

\section{Example Inputs and Outputs}
\label{appendix:example_inputs_and_outputs}
The following prompts were used to produce example outputs. The model used was Llama 3.3 70B, and the research strategy employed was the minimal strategy given in \Cref{appendix:minimal_research_strategy}.

\subsection{African or European swallow?}
\subsubsection{Input}
\begin{lstlisting}[breaklines]
What is the airspeed velocity of an unladen swallow? Produce a detailed report on the subject, including the occurences of the information in the popular culture. Condense your output into three sections.
\end{lstlisting}

\subsubsection{Raw markdown output}
\begin{lstlisting}[breaklines]
# The Airspeed Velocity of an Unladen Swallow: A Cultural and Technical Exploration

## Origins and Popular Culture Significance
The question of the airspeed velocity of an unladen swallow originates from the 1975 film *Monty Python and the Holy Grail* [[0,1]]. In the film, King Arthur engages in a humorous debate with a castle guard and later a troll, discussing the feasibility of coconut transportation by swallows and the specifics of swallow airspeed [[0]]. The dialogue's absurdity, particularly the distinction between African and European swallows, has become a cultural reference point for overly trivial or technical inquiries [[0,1]]. This reference has permeated various media, including technology, with virtual assistants like Siri and computational engines such as Wolfram Alpha providing humorous and scientific responses, respectively [[0]].

## Technical Analysis and Accuracy
From a technical standpoint, the airspeed velocity of an unladen swallow has been subject to analysis and myth-busting [[1]]. Contrary to the film's suggestions, the average European swallow weighs less than an ounce and flaps its wings approximately 12 times per second [[1]]. A study cited in *Interesting Engineering* estimates the airspeed velocity to be around 20.1 miles per hour [[1]], while another analysis by Jonathan Corum in 2003 suggests an average cruising airspeed of roughly 11 meters per second or 24 miles per hour [[0]]. These findings highlight the discrepancy between the film's comedic portrayal and real-world ornithological data.

## Enduring Cultural Impact
The reference to the airspeed velocity of an unladen swallow has evolved into a lasting symbol of absurdity and intellectual curiosity in popular culture [[0,1]]. It is frequently used in memes, videos, and online discussions to humorously critique trivial debates or pointless inquiries [[0]]. The phrase's enduring popularity across different platforms and communities, from computing to social media, underscores its status as a cultural touchstone, blending humor with a touch of intellectualism characteristic of Monty Python's influence [[0,1]].

---
 - [[0]] [What Is the Airspeed Velocity of an Unladen Swallow?][0]
 - [[1]] [Monty Python and the Holy Grail's biggest questions, answered - SYFY][1]


[0]: https://knowyourmeme.com/memes/what-is-the-airspeed-velocity-of-an-unladen-swallow
[1]: https://www.syfy.com/syfy-wire/monty-python-and-the-holy-grail-swallows-coconuts-rabbits-black-knight
\end{lstlisting}

\subsection{Significant events on a chosen day}
\subsubsection{Input}
\begin{lstlisting}[breaklines]
Produce a report on the most significant events that occurred on the 1st of May 2025. Write 3 sections.
\end{lstlisting}

\subsubsection{Raw markdown output}
\begin{lstlisting}
# Report on the Most Significant Events of May 1st, 2025

## 1. International Labour Day (May Day)
May 1st, 2025, marks the global observance of **International Labour Day**, also known as **May Day** [[0,1,2,3,4]]. This day commemorates the labour movement and the fight for workers' rights, advocating for fair wages, safe working conditions, and social justice [[2,4]]. The origins of the day date back to the late 19th century, specifically the **Haymarket Affair** in Chicago in 1886, which drew international attention to workers' demands for an eight-hour workday [[2]]. In 2025, the day continues to emphasize the importance of protecting workers' rights and promoting fair treatment across industries and nations [[2,4]]. It is celebrated worldwide with parades, rallies, and events organized by trade unions [[2]].

## 2. Maharashtra Day and Gujarat Day
In India, May 1st also coincides with **Maharashtra Day** and **Gujarat Day**, commemorating the formation of the states of **Maharashtra** and **Gujarat** on **1 May 1960** [[0,3,4]]. Both states observe a **state holiday** and organize parades, cultural programs, and ceremonies to showcase their heritage and progress [[0,3,4]]. These regional celebrations highlight the cultural, economic, and historical significance of Maharashtra and Gujarat, known for their vibrant traditions and contributions to India's development [[4]].

## 3. Global Significance and Reflection
The significance of May 1st, 2025, lies in its dual role as a day of global solidarity for workers' rights and a celebration of regional identity and heritage [[4]]. While the provided information does not specify unique events exclusive to May 1st, 2025, the day's importance is rooted in these established observances [[4]]. International Labour Day and the regional celebrations in India serve as a reminder of the ongoing struggle for worker dignity and the promotion of social and economic justice worldwide [[2,4]].

---
 - [[0]] [Important Days in May 2025, List of National and International Events][0]
 - [[1]] [Important Dates in May 2025: Global Events, Holidays][1]
 - [[2]] [Important National and International Days in May 2025: Themes ...][2]
 - [[3]] [Important Days in May 2025: National and International Dates List in May][3]
 - [[4]] [May 2025 Calendar: Full List of Important National & International Days ...][4]


[0]: https://currentaffairs.adda247.com/important-days-in-may/
[1]: https://importantdayshub.com/important-dates-in-may-2025/
[2]: https://www.socialworkin.com/2025/04/important-national-international-days-may-2025.html
[3]: https://www.jagranjosh.com/general-knowledge/important-days-in-may-1557143140-1
[4]: https://dailyschoolassembly.com/may-important-days/
\end{lstlisting}

\subsection{Stock Movements on a Chosen Day}
\subsubsection{Input}
\begin{lstlisting}[breaklines]
Produce a detailed report on the US stock movements on the Thursday 24th of April 2025. Note the opening and closing prices. In terms of formatting, make three sections: one focusing on the conditions at which stocks opened, one focusing on the conditions under which the stock closed, and one putting the daily movements into a wider perspective.
\end{lstlisting}

\subsubsection{Raw markdown output}
Any hallucinations are not due to incorrect search results but to the incorrect comprehension of the search results by the language model.
\begin{lstlisting}[breaklines]
# US Stock Movements Report - April 24, 2025

## Opening Conditions
The US stock market opened with mixed signals across key indices. The S&P 500 opened at 5395.92 [[0]], while the Dow Jones Industrial Average (DJIA) started the day at 39815.01 [[0]]. The NASDAQ Composite began trading at 16,880.29 [[0]]. These opening prices set the stage for a day that would see significant movement, influenced by ongoing tariff negotiations and rate cut expectations [[1]]. Notably, the SPY opened at $540.43, reflecting the broader market sentiment [[1]].

## Closing Conditions
By the close of trading, the S&P 500 had risen to 5375.86, marking a 1.67% increase from the previous close [[0]]. The DJIA closed at 39606.57, up 1.07% from its previous close [[0]], with a points change of 419.58 [[2]]. The NASDAQ Composite ended the day at 16,708.05, achieving a 2.5% gain [[0]]. The SPY closed strongly at $546.48, breaking through key resistance levels and signaling continued momentum [[1]]. The S&P 500's close at 5,375.86 represented its highest level since April 15 [[2]].

## Wider Perspective
The day's movements were part of a broader three-day rally, with the NASDAQ, S&P 500, and DJIA gaining more than 8%, 6%, and 5%, respectively, over this period [[1]]. Optimism around tariff negotiations with India and South Korea, coupled with dovish comments from the Cleveland Fed President, contributed to the bullish sentiment [[1]]. The market also responded positively to strong earnings reports, such as Alphabet's better-than-expected results and its announcement of a stock buyback and dividend hike [[1]]. Technical indicators presented a mixed outlook, with the RSI suggesting neutral momentum for the S&P 500 and the MACD indicating bearish trends for the DJIA and NASDAQ [[0]]. The S&P 500's exit from correction territory further highlighted the market's resilience, with the index rising above the 5481.05 level needed to mark a 10% recovery from its recent low [[3]]. However, some sources noted a lack of specific data for April 24, 2025, in historical records [[4]].

---
 - [[0]] [Global stock market indices: How the world markets performed today - 24 ...][0]
 - [[1]] [Market Insights: Thursday, April 24th, 2025][1]
 - [[2]] [Stock Market News for Apr 24, 2025 - Yahoo Finance][2]
 - [[3]] [Stock Market News From April 24, 2025: Dow, S&P 500, Nasdaq ... - Barron's][3]
 - [[4]] [S&P 500 Historical Data (SPX) - Investing.com][4]


[0]: https://www.editorji.com/business-news/markets/global-stock-market-indices-performed-today-april-24-2025-1745491687078
[1]: https://www.monetaritrading.com/blog/market-insights-thursday-april-24th-2025
[2]: https://finance.yahoo.com/news/stock-market-news-apr-24-135000428.html
[3]: https://www.barrons.com/livecoverage/stock-market-today-042425
[4]: https://www.investing.com/indices/us-spx-500-historical-data
\end{lstlisting}

\subsection{Ulysses Grant}
\subsubsection{Input}
\begin{lstlisting}[breaklines]
Produce a research report on General Ulysses S. Grant, focusing on his Civil War military leadership, key battles, strategic approach, and his influence during Reconstruction as president. The report should be structured into five sections: introduction, military career, leadership style, political legacy, and conclusion. Use at least three scholarly sources, include citations, and format the report with clear section headings.
\end{lstlisting}

\subsubsection{Raw markdown output}
\begin{lstlisting}[breaklines]
# Ulysses S. Grant: A Study in Military Leadership and Political Legacy

## Introduction
Ulysses S. Grant, born Hiram Ulysses Grant, rose from a mediocre military career to become one of the most renowned generals in U.S. history, credited with winning the Civil War and preserving the Union [[4]]. His leadership during the war and subsequent presidency shaped the nation's trajectory during Reconstruction. This report examines Grant's military leadership, key battles, strategic approach, and his influence during Reconstruction as president.

## Military Career
Grant's military career began with his appointment as colonel of the 21st Illinois Infantry in 1861, quickly rising to brigadier general [[4]]. His early victories at Fort Donelson and Fort Henry in 1862 earned him the nickname "Unconditional Surrender" Grant [[4]]. Despite facing criticism and calls for his removal after the Battle of Shiloh, Lincoln supported Grant, recognizing his tenacity [[4]]. Grant's strategic masterpiece at Vicksburg in 1863 split the Confederacy and marked a turning point in the war [[2], [3], [4]]. His subsequent successes at Chattanooga and his appointment as general-in-chief of all Union armies in 1864 solidified his position as a national hero [[2], [3], [4]].

## Leadership Style
Grant's leadership style was characterized by his determination and willingness to take risks. He emphasized the importance of relentless pressure on Confederate forces, stating, "The art of war is simple enough. Find out where your enemy is. Get at him as soon as you can. Strike him as hard as you can, and keep moving on" [[1]]. This approach was evident in his Overland Campaign, where he suffered heavy casualties but ultimately wore down Lee's army [[2], [3]]. Grant's ability to adapt and his strategic acumen, learned from mentors like Zachary Taylor and Winfield Scott, were crucial to his success [[3]]. His decision to cut supply lines during the Vicksburg campaign demonstrated his willingness to challenge conventional military tactics [[3]].

## Political Legacy
Grant's presidency, from 1869 to 1877, focused on preserving the Union and enforcing Reconstruction policies [[2]]. He supported the Fifteenth Amendment, protecting African American voting rights, and deployed federal troops to combat the Ku Klux Klan [[2]]. Despite scandals tarnishing his administration, Grant's commitment to Reconstruction and civil rights underscored his dedication to a unified nation [[2], [4]]. His memoirs, written after leaving office, provided valuable insights into his military strategies and personal experiences [[4]].

## Conclusion
Ulysses S. Grant's military leadership and strategic brilliance were instrumental in the Union's victory in the Civil War [[2], [3], [4]]. His presidency, though marred by controversy, played a crucial role in shaping Reconstruction and protecting the rights of formerly enslaved individuals [[2], [4]]. As a national hero and a symbol of Union preservation, Grant's legacy endures as a testament to his unwavering resolve and military acumen [[1], [2], [3], [4]].

---
 - [[0]] [List of the Battles Ulysses S. Grant Fought In - Ranker][0]
 - [[1]] [Ulysses S. Grant: A Map of Key Civil War Battles | HISTORY][1]
 - [[2]] [Ulysses S. Grant and the American Civil War - Wikipedia][2]
 - [[3]] [Grant's Greatest Battles | American Experience | PBS][3]
 - [[4]] [Ulysses S. Grant - American Battlefield Trust][4]


[0]: https://www.ranker.com/list/list-of-all-ulysses-s-grant-battles/reference
[1]: https://www.history.com/shows/grant/interactives/ulysses-s-grant-battle-map
[2]: https://en.wikipedia.org/wiki/Ulysses_S._Grant_and_the_American_Civil_War
[3]: https://www.pbs.org/wgbh/americanexperience/features/grants-greatest-battles/
[4]: https://www.battlefields.org/learn/biographies/ulysses-s-grant
\end{lstlisting}

\end{document}